\documentclass[review]{elsarticle}
\graphicspath{ {./figures/} }
\usepackage{hyperref}
\usepackage{float}
\usepackage{verbatim} %comments
\usepackage{apalike}
\restylefloat{figure}
\restylefloat{table}

\usepackage{graphicx}
\usepackage{multirow}
\usepackage{caption} 
\usepackage{makecell}

\journal{Expert Systems with Applications}

\begin{document}
\begin{frontmatter}

\begin{titlepage}
\begin{center}
\vspace*{1cm}

\textbf{ \large Sarcasm Detection Framework Using Context, Emotion and Sentiment Features}

\vspace{1.5cm}

% Author names and affiliations
Oxana Vitman$^{a}$ (oksana.vittmann@gmail.com), Yevhen Kostiuk$^a$ (kosteugeneo@gmail.com), Grigori Sidorov$^a$ (sidorov@cic.ipn.mx), Alexander Gelbukh$^a$ (gelbukh@cic.ipn.mx) \\

\hspace{10pt}

\begin{flushleft}
\small  
$^a$ Instituto Polit\'ecnico Nacional, Centro de Investigaci\'{o}n en Computaci\'{o}n, Av. Juan de Dios Batiz, s/n, 07320, Mexico City, Mexico\\
\begin{comment}
Clearly indicate who will handle correspondence at all stages of refereeing and publication, also post-publication. Ensure that phone numbers (with country and area code) are provided in addition to the e-mail address and the complete postal address. Contact details must be kept up to date by the corresponding author.
\end{comment}

\vspace{1cm}
\textbf{Corresponding Author:} \\
Alexander Gelbukh \\
Instituto Polit\'ecnico Nacional, Centro de Investigaci\'{o}n en Computaci\'{o}n, Av. Juan de Dios Batiz, s/n, 07320, Mexico City, Mexico \\
Tel: (559) 188-7293 \\
Email: gelbukh@cic.ipn.mx

\end{flushleft}        
\end{center}
\end{titlepage}

\title{Sarcasm Detection Framework Using Context, Emotion and Sentiment Features}

\author[label1]{Oxana Vitman}
\ead{oksana.vittmann@gmail.com}

\author[label1]{Yevhen Kostiuk}
\ead{KostEugeneO@gmail.com}

\author[label1]{Grigori Sidorov}
\ead{sidorov@cic.ipn.mx}

\author[label1]{Alexander Gelbukh \corref{cor1}}
\ead{gelbukh@cic.ipn.mx}

\cortext[cor1]{Alexander Gelbukh.}
\address[label1]{Instituto Polit\'ecnico Nacional, Centro de Investigaci\'{o}n en Computaci\'{o}n, Av. Juan de Dios Batiz, s/n, 07320, Mexico City, Mexico}

\begin{abstract}
Sarcasm detection is an essential task that can help identify the actual sentiment in user-generated data, such as discussion forums or tweets. Sarcasm is a sophisticated form of linguistic expression because its surface meaning usually contradicts its inner, deeper meaning. Such incongruity is the essential component of sarcasm, however, it makes sarcasm detection quite a challenging task. In this paper, we propose a model, that incorporates different features to capture the incongruity intrinsic to sarcasm. We use a pre-trained transformer and CNN to capture context features, and we use transformers pre-trained on emotions detection and sentiment analysis tasks. Our approach outperformed previous state-of-the-art results on four datasets from social networking platforms and online media.
\end{abstract}

\begin{keyword}
sarcasm detection \sep context \sep emotion \sep sentiment \sep transformers
\end{keyword}

\end{frontmatter}

\section{Introduction}
\label{introduction}

The world's use of online communications is expanding quickly since social media has become the most important source of news and public opinion about almost every daily topic.
The most popular application of social media analysis is detecting consumer sentiment to help businesses and online merchants to address the needs of their clients, including handling and resolving complaints. 

However, not always emotions or sentiments are expressed directly. Frequently, social media users tend to make their posts or messages sarcastic in order to get better responses from other users and stimulate the virality of social media content. Moreover, negative sarcastic tweets attract substantially higher social media responses when compared to actual negative tweets~\citep{peng2019discovering}. Thus, sarcasm identification in online communications, discussion forums, and e-commerce websites has become essential for fake news detection, sentiment analysis, opinion mining, and hate speech detection~\citep{majumder2019sentiment, reyes2022linguistic}. 

 For example, applying sentiment analysis task for feedback classification,  we might capture only surface meaning for those feedbacks containing sarcasm and classify it as a positive one. However, by applying sarcasm detection model to the same feedback, and detecting sarcasm, we should change the label from positive to negative, since sarcasm is a negative sentiment hidden beneath a positive surface~\citep{riloff2013sarcasm}.
% However, sarcasm detection appears to be quite a challenging task due to its sophisticated nature.

 % The surface meaning of sarcasm frequently contrasts with the underlying deeper meaning, making it a particularly complex kind of verbal expression. Although 
 
 This incongruity is the essential component of sarcasm, the intention may also be to appear humorous, make fun of someone, or show contempt. As a result, sarcasm is seen as an extremely sophisticated and intelligent language construct that presents a number of difficulties in the perception of emotions.

The following example of a tweet illustrates the above-mentioned nuances: \emph{``On our 6 am walk, my daughter asked me where the moon goes each morning. I let her know it’s in heaven, visiting daddy’s freedom''}.

On the surface, this statement seems to indicate that the speaker is enjoying his morning walk with his daughter and telling her amusing stories. However, a close examination of the speaker's emotions and sentiment reveals that the speaker is unhappy and experiencing some unpleasant emotions at the time of speaking.

We hypothesize that detecting polarity between the surface and true sentiment is an important component of sarcasm detection. We propose a model that utilizes contextual features, as well as sentiment and emotions features to learn their dependencies related to sarcasm. 

% Moreover, it makes sarcasm detection an essential component for such tasks as sentiment analysis, hate speech detection, and opinion mining. For example, applying sentiment analysis task for feedback classification, for those feedbacks which contains sarcasm we might capture only surface meaning and classify it as a positive one. However, by applying sarcasm detection model to the same feedback, and detecting sarcasm, we should change the label from positive to negative, since sarcasm is a negative sentiment hidden beneath a positive surface~\citep{riloff2013sarcasm}.
% This is where incongruity between sentiment and emotions plays an important role. Sentiment, emotion, and sarcasm are highly interconnected, and one helps in the understanding of the others.
% We propose a model, which utilizes sentiment analysis and emotion detection and learns their dependencies related to sarcasm. We hypothesized that learning a pattern of contradiction between surface sentiment and intended sentiment is a key component in sarcasm detection.

\section{Related Work}
While sarcasm detection is still a relatively new field, it has recently drawn more attention from researchers due to the rapid growth of social media and the need for sentiment analysis therein. A number of approaches have been developed and studied in this regard. In this section, we are going to pay attention to works implementing transformers, neural networks, or focusing on a relation between context, sentiment, emotion, and sarcasm. 

The research~\cite{math10050844} proposed a model based on BERT~\citep{10.5555/3295222.3295349} fine-tuned on related intermediate tasks, such as sentiment classification and emotion detection, before fine-tuning it on the target task. Experimental results on three datasets that have different characteristics showed that the intermediate task transfer learning model outperforms many previous models.

The study~\cite{babanejad-etal-2020-affective}  proposed two novel deep neural network models for sarcasm detection. Models add to the BERT architecture two components, namely affective feature embedding and contextual feature embedding. The two models are different in the way the two components are combined and the input to the affective feature embedding component.

 The paper~\cite{ghosh2016fracking} proposed a neural network semantic model composed of Convolution Neural Network (CNN), followed by a Long Short Term Memory (LSTM) network and a Deep Neural Network (DNN).  The proposed model outperforms Support Vector Machine (SVM) applied to the same dataset. 

 In the work~\cite{akula2021interpretable}, the authors focus on detecting sarcasm in textual conversation from social media websites and online forums.  They developed a model consisting of a multi-head self-attention module and gated recurrent units. The multi-head self-attention module helps in identifying important sarcastic keywords from the input, and the recurrent units learn long-range dependencies between these keywords to classify the input text more accurately.  The proposed approach achieved state-of-the-art results on multiple datasets from social media platforms and online networks.

In the paper~\cite{chauhan2020sentiment}, authors have hypothesized that sarcasm is closely related to sentiment and emotion. They proposed two attention mechanisms, where the first segment learns the relationship between the different parts of the sentence, and the second segment focuses on the same part of the sentence across the modalities. Finally, representations from both attentions are merged for a multi-class classification task. The research proved that emotions and sentiment help to improve the effect of satire detection, however, the experiments were performed on MUStArd dataset~\cite{mustard}, which consists of audiovisual utterances accompanied by its context, thus providing additional cues for sarcasm detection task. 

\section{Description of Datasets}

We conducted experiments on four benchmark datasets: two Reddit~\citep{khodak-etal-2018-large} subreddits datasets: SARC/movies and SARC/technology, a subset of Internet Argument Corpus-V2~\citep{walker2012corpus} (IAC-V2), and Twitter~\citep{ptacek-etal-2014-sarcasm}. All of them have been widely used in evaluating sarcasm detection. The details are shown in Table~\ref{tab:datasets}.

\begin{table}[!htb]
\caption{Statistics the experimental data.}
\label{tab:datasets}
% \begin{tabular}{|>{\centering\arraybackslash}|>{\cenon\arraybackslash} |>{\centering\arraybackslash} |>{\centering\arraybackslash}  |>{\centering\arraybackslash} |}
\begin{tabular}{|c|c|c|c|c|}
\hline & \multicolumn{2}{c|}{Train}  & \multicolumn{2}{c|}{Test}\\ \cline{2-5} 
% \multirow{Datasets} & \multicolumn{1}{c|}{Sarcastic} & {Non-sarcastic} & \multicolumn{1}{c|}{Sarcastic} & {Non-sarcastic} \\
\multirow{-2}{*}{Datasets} & {Sarcastic} & {Non-sarcastic} & {Sarcastic} & {Non-sarcastic} \\ 
\hline
SARC/movies & {2,533}      & 2,707    & {641} & 669           \\
SARC/technology  & {2,738}      & 1,815     & {677}       & 462           \\
IAC\_V2    & {2,616}      & 2,600          & {644}       & 660           \\
\makecell{Twitter \\ (Ptá\v cek et al., 2014)} & {22,323}     & 25,785         & {5,648}      & 6,379        \\ \hline
\end{tabular}
\end{table}

\subsection{Reddit}
\cite{khodak-etal-2018-large} collected SARC, a corpus comprising of 600,000 sarcastic comments on Reddit. We used two subreddits: SARC/movies and SARC/technology for sarcasm detection. The subreddits contain news and discussions concerning films and development, application, and related issues of technology, respectively.

\subsection{Internet Argument Corpus}
Internet Argument Corpus (IAC)~\citep{walker2012corpus} was collected from online political debates forum. IAC-V2~\citep{abbott2016internet} is the subset of the Internet Argument Corpus.
%  Each instance is annotated with a label, either ``sarcasm'' or ``nonsarcasm''. Compared to tweets, texts of IAC-V2 are much longer and more normative. 
 IAC-V2 divides sarcasm into three sub-types, (i.e., general sarcasm, hyperbole, and rhetorical questions). We used the largest subset (general sarcasm) in our experiments.
 
\subsection{Twitter}
We used Twitter dataset provided by Tomá\v s  Ptá\v cek ~\citep{ptacek-etal-2014-sarcasm}, who collected a self-annotated corpus of tweets with the \emph{\#sarcasm} hashtag. We used the English balanced version.

\section{Proposed Method}
\subsection{Main Idea}

The main idea of the architecture is to obtain text representation in a different ways. Thus, the model consists of four different blocks, and each one of them derives different text features.

\begin{figure}[!htb]
\centering
\includegraphics[width  = \textwidth]{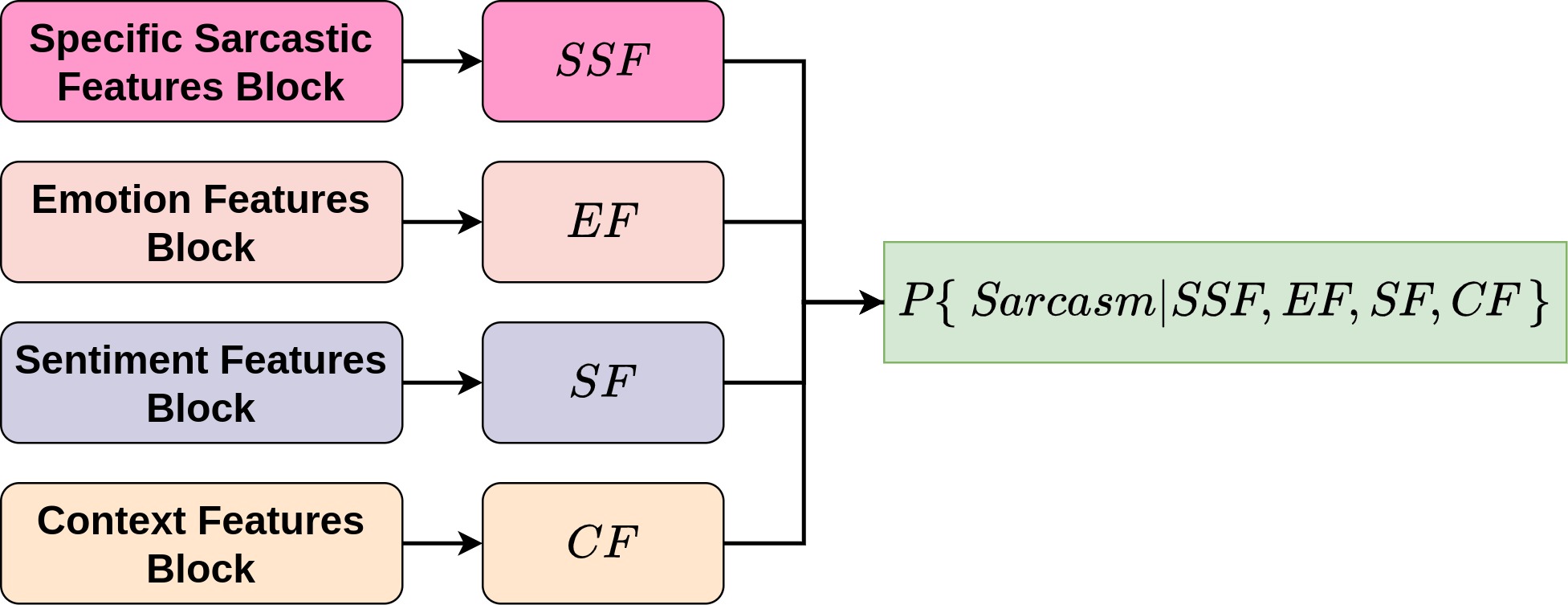}
\caption{Scheme of the proposed architecture.}
\label{fig:main_idea}
\end{figure}

 The first block is a RoBERTa-based~\citep{liu2019roberta}  model which was pre-trained on merged train chunks of datasets, thus it learned specific sarcasm-related features of the text. The second and third blocks are transformers pre-trained on emotion detection and sentiment analysis tasks respectively. The outputs of these blocks are text representations with regard to emotion and sentiment features. Finally, the fourth block, based on CNN architecture, extracts contextual features and semantic relationships of the words.  All text representations are concatenated and the output layer of the model predicts whether input text is sarcastic or not, based on all the features that were passed to it (see Figure~\ref{fig:main_idea}).
 
\subsection{Overview of the Proposed Approach}
The model consists of four blocks: Sarcasm Pre-Trained Transformer (SarcPTT), Emotion Detection Pre-Trained Transformer (EmoDPTT), Sentiment Analysis Pre-Trained Transformer (SentAPTT), and CNN block. EmoDPTT and SentAPTT are used as feature extractors and not trainable during model fitting. Other modules (CNN and SarcPTT) are trainable. A detailed overview of the blocks is presented in the following subsections. 

\begin{figure}[!htb]
\centering
\includegraphics[width=\textwidth]{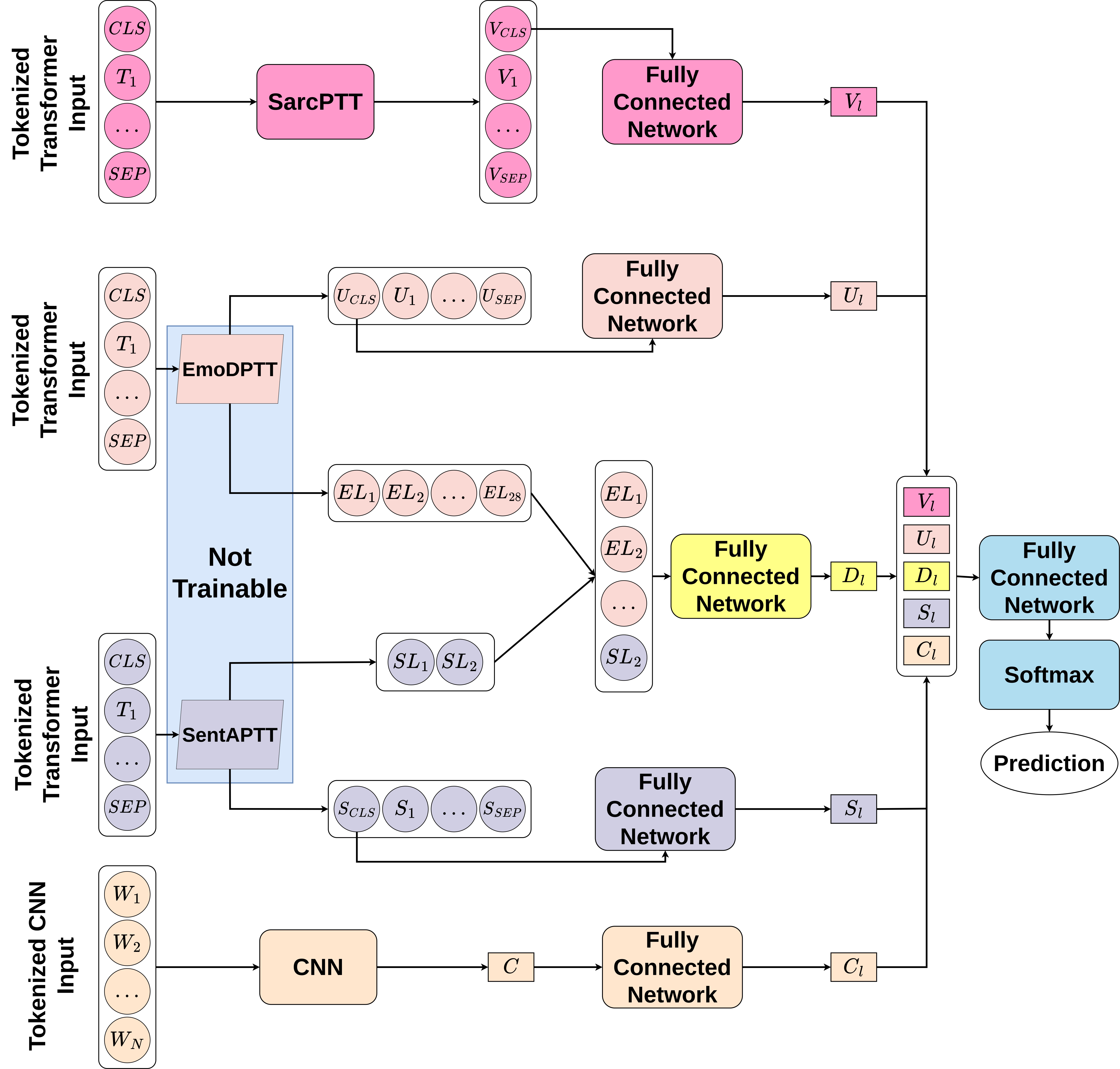}
\caption{Proposed architecture of the model.}
\label{fig:arch}
\end{figure}

The general flow is presented in Figure~\ref{fig:arch}. The input text is tokenized via transformer tokenizer into tokens $\{CLS, T_1, ..., SEP\}$ and passed through SarcPTT. The output of this step is last hidden state, but only vector representation $V_{CLS}$ of the first token ($CLS$) is used for further processing. After that, the output is passed through Fully Connected Network, and new vector $V_l$ is obtained.

The non-trainable blocks (EmoDPTT and SentAPTT) are used as feature extractors. The tokenized text ($\{CLS_{e}, T_1^e, ..., SEP_e\}$ and $\{CLS_{s}, T_1^s, ..., SEP_s\}$ respectively) is passed through the models, the following features are obtained: 
\begin{itemize}
    \item EmoDPTT vector representation of the first token of the last hidden state: $U_{CLS}$;
    \item Probability distribution  of dimension $28$ of EmoDPTT labels, e.g., amusement, relief, disgust, neutral, etc.: $\{EL_1, EL_2, ..., EL_{28} \}$;
    \item SentDATT vector representation of the first token of last hidden state: $S_{CLS}$;
    \item Probability distribution of dimension $2$ of SentAPTT labels, i.e., positive and negative sentiments $\{SL_1, SL_2\}$.
\end{itemize}

Similarly to the SarcPTT output, text representations are passed through Fully Connected Networks, and new vectors are obtained: $S_l$ and $U_l$. Probability distributions of labels are concatenated and passed through a Fully Connected Network, producing vector $D_l$ as output.

% For passing data through CNN block, the input text was embedded, and text representations with dimension $(N, d)$ were obtained, where $N$ is a number of tokens in the texts, and $d$ is embedding dimension.
For passing data through CNN block, the input text $T = \{W_1, W_2, ..., W_N \}$ is tokenized using $nltk$ library\footnote{\url{https://www.nltk.org/}},  where each $W_i$ represents a tokenized word. The text $T$ is passed through CNN, and text representations are obtained as vector $C$. Vector $C$ is passed through a Fully Connected Network, transforming it into vector $C_l$.

As a final step, all the output feature vectors $V_l, U_l, D_l, S_l, C_l$ are concatenated and passed through a Fully Connected Network and Softmax transformation, obtaining the prediction.

We made our code available at GitHub\footnote{\url{https://github.com/Wittmann9/SarcasmTuneUntrainableSentEmo}}.

% \begin{figure}[htb!]
% \centering
% \includegraphics[width=\textwidth]{SarcasmModel-Copy of Copy of Copy of Page-2.png}
% \caption{Proposed architecture of the model.}
% \label{fig:arch}
% \end{figure}

% We have implemented[integrated?] several Neural models[modules] in order to obtain different vector representations of our texts. Particularly, we passed out datasets through pretrained RoBERTa, BERT pretrained on emotion classification task\footnote{https://huggingface.co/bhadresh-savani/bert-base-go-emotion}, RoBERTa pretrained on sentiment classification task\footnote{https://huggingface.co/siebert/sentiment-roberta-large-english}, CNN model. We have concatenated all obtained vectors representations and processed it through linear classifier and softmax function to compute final predictions. [We hypothesise t]The main idea of this approach is to emphasize different [sentiment]? signals and capture [information?]incongruity intrinsic to sarcastic speech.

% Есть эмоушен, есть сентимен, очевидно, что онимежду собой связнеы, anger == negative sentiment, но и сарказм постороен очень особо., где между эмоушен и сентимент может воникать противоречие, также сарказм очень отличается отобычного языка, соотв наша гипотеза лежит в том чтораспределение эмоушен и сентимент тоже будет отличаться от обычного языка

\subsection{Sarcasm Pre-Trained Transformer}
We merged all training parts of the datasets and then we pre-trained RoBERTa model on it using masked language modeling objective. 
% For each dataset we then passed tokenized texts through pretrained RoBERTa model and obtained CLS token representation of last hidden state. We passed this representations through linear model. 

\subsection{Emotion Detection Pre-Trained Transformer}
% We passed our datasets through BERT pretrained on emotion classification task and obtaining  corresponding last hidden CLS token representations and labels distribution.
As an emotion feature extractor, we used BERT model\footnote{\url{https://huggingface.co/bhadresh-savani/bert-base-go-emotion}} pre-trained on GoEmotions dataset~\citep{demszky2020goemotions} on multi-label (28) emotion classification task.
The input text was passed through this model and corresponding vector representation of CLS token of the last hidden state was obtained, as well as labels' distribution.

\subsection{Sentiment Analysis Pre-Trained Transformer}
% We also passed our datasets through RoBERTa pretrained on sentiment classification task, obtaining  corresponding last hidden CLS token representations and labels distribution.
As a sentiment feature extractor, we used SiEBERT model~\citep{hartmann2022}. It is a fine-tuned on sentiment classification task RoBERTa-large model~\citep{liu2019roberta}. 
The model was trained on 15 datasets from different text sources (reviews, tweets, etc.). 
We passed the input text through the model and corresponding vector representation of CLS token of the last hidden state was obtained, as well as labels' distribution.

\subsection{CNN}
\begin{figure}[!htb]
\centering
\includegraphics[width  = \textwidth]{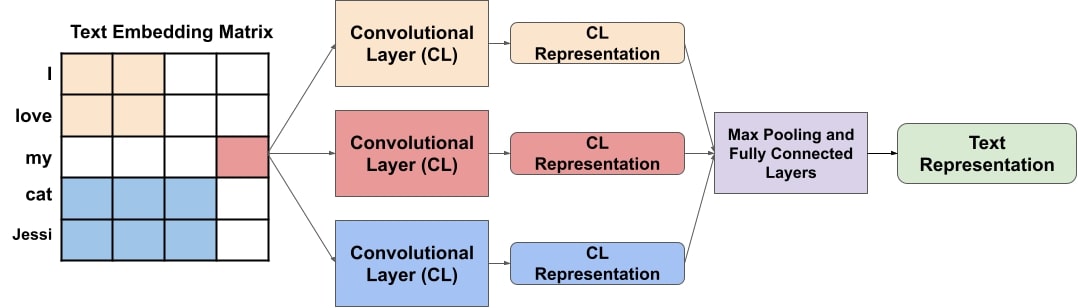}
\caption{CNN block.}
\label{fig:cnn}
\end{figure}

We used the CNN architecture following~\cite{kimCNN} (see Figure~\ref{fig:cnn}). For processing the input text through CNN, we built a vocabulary dictionary for each dataset separately to create an embedding layer. We used Glove Common Crawl\footnote{\url{https://nlp.stanford.edu/projects/glove/}} pre-trained vectors (42 billion words version). For words with no pre-trained vectors, we checked their stemmed versions. If no pre-trained embedding was found, a random vector was initialized. Then, the input text was encoded into the embedding matrix of shape $(N_W, 300)$, where $N_W$ is the number of words in the input text and 300 is an embedding dimension.

We then used convolutions with different filter sizes to extract feature maps from the embedding matrix. Next, we applied the ReLU activation and max-over-time-pooling to reduce each feature map to a single scalar.
Then we concatenated these scalars into a vector and obtained vector representations for the texts.

% The idea here is that each filter captures different semantic signals in the sentence and max-pooling records only the strongest signal over the sentence.

\section{Description of Baselines}

Since prior studies lack uniform datasets, we run baseline comparison experiments using our datasets and either open source code or reproduced code.

\subsection{NBOW}
Neural bag-of-words (NBOW) baseline takes an average of word vectors in the given text as sentence representation and feeds it into a standard logistic regression model. As word vector representations we used GloVe 100-dimensional vectors~\citep{pennington2014glove} pre-trained on Wikipedia and an archive of English newswire text named Gigaword 5, version containing 6 billion tokens. 

\subsection{CNN}
We used CNN configuration following~\cite{kimCNN}.
The overall architecture is as follows. The padded embedded sentences were processed via the CNN cells. Next, the ReLU activation function and Max Pooling were applied. The concatenated outputs from the previous step were processed by linear layers to produce the distribution of the classes.

\subsection{CNN-LSTM-DNN}
This model~\citep{ghosh2016fracking} is a combination of CNN, LSTM, and DNN. It stacks two layers of convolution and two LSTM layers, then passes the output to a DNN for prediction.

\subsection{SAWS}
SAWS~\citep{pan2020modeling} model is a self-attention mechanism of weighted sentence fragments. It models the incongruity between sentence fragments.

\subsection{ELMo}
This model~\citep{ilic-etal-2018-deep} uses character-level vector representations of words, based on embeddings from ELMo~\citep{peters-etal-2018-deep} language model architecture. Subsequently, word embeddings were passed on to a BiLSTM, and the output hidden states were max-pooled and fed to the 2-layer feed-forward network. The output of this step was then fed to the final layer of the model, which performed binary classification.

\section{Experiments}
We implemented our model using Pytorch Lightning~\citep{falcon2019pytorch} and transformers library~\citep{wolf-etal-2020-transformers}. For each dataset, we experimented with different sets of hyperparameters. The best-performing hyperparameters are presented in Table~\ref{tab:hyperparams}.

\begin{table}[!htb]
\caption{Best performing hyperparameters for various datasets.}
\label{tab:hyperparams}
% \begin{tabular}{{|>{\centering\arraybackslash}m{2cm}|>{\centering\arraybackslash} m{2cm}|>{\centering\arraybackslash} m{2cm} |>{\centering\arraybackslash} m{2cm} |>{\centering\arraybackslash} m{2cm} |}}
\begin{tabular}{|c|c|c|c|>{\centering\arraybackslash} m{3cm}|}
\hline& \multicolumn{4}{c|}{Datasets} \\ \cline{2-5} 
{Parameter} & {SARC/movies} & {SARC/tech} & {IAC\_V2} & Twitter (Ptá\v cek et al., 2014) \\ \hline
max\_length                         & {18}                & {14}              & \multicolumn{1}{c|}{16}                                & 16             \\
max\_epochs                       & \multicolumn{1}{c|}{12}                & \multicolumn{1}{c|}{30}              & \multicolumn{1}{c|}{20}                                & 20             \\
lr                                & \multicolumn{1}{c|}{1e(-5)}            & \multicolumn{1}{c|}{1e(-5)}          & \multicolumn{1}{c|}{1e(-5)}                          & 1e(-5)         \\
batch\_size                       & \multicolumn{1}{c|}{8}                 & \multicolumn{1}{c|}{4}               & \multicolumn{1}{c|}{32}                                & 32             \\ \hline
\end{tabular}
\end{table}

\section{Results}

We used precision, recall, macro-weighted F1, and accuracy scores as the evaluation metrics. We emphasized F1 score as the main metric in our analysis. Table~\ref{tab:results} reports the results of our method as well as baselines. 

\begin{table}[!htb]
\caption{Results of the experiments.}
\label{tab:results}
\begin{tabular}{|>{\centering\arraybackslash}m{3cm}|c|c|c|c|c|}
\hline
Datasets                                       & Model        & Acc.           & Precision & Recall & F1    \\ \hline
\multirow{6}{*}{SARC/movies}                    & CNN-LSTM-DNN & 58.78          & 57.85     & 58.03  & 57.94 \\
                                               & NBOW         & 59.69          & 59.71     & 59.69  & 59.69 \\
                                               & CNN          & 64.73          & 64.77     & 64.77  & 64.73 \\
                                               & SAWS         & 65.35          & 65.42     & 65.23  & 65.19 \\
                                               & ELMo         & 68.01          & 71.60      & 57.41  & 63.72 \\
                                               & Our model    & 73.36          & 72.05     & 74.41  & \textbf{73.21} \\ \hline
\multirow{6}{*}{SARC/technology}              & CNN-LSTM-DNN & 56.10           & 64.58     & 57.90   & 61.06 \\
                                               & NBOW         & 64.27          & 63.3      & 64.27  & 64.27 \\
                                               & CNN          & 66.37          & 66.79     & 67.38  & 66.19 \\
                                               & SAWS         & 66.9           & 65.52     & 65.11  & 65.25 \\
                                               & ELMo         & 72.96          & 75.59     & 80.50   & 77.97 \\
                                               & Our model    & 75.24          & 76.73     & 83.75  & \textbf{80.08} \\ \hline
\multirow{6}{*}{IAC\_V2}                       & CNN-LSTM-DNN & 65.49          & 62.63     & 74.69  & 68.13 \\
                                                & NBOW         & 72.24          & 72.17     & 71.27  & 71.72 \\
                                               & CNN          & 70.46          & 71.41     & 70.60   & 70.23 \\
                                               & SAWS         & 74.62          & 74.62     & 74.61  & 74.61 \\
                                               & ELMo         & 76.83 & 78.07     & 73.55  & 75.75 \\
                                               & Our model    & 85.12 & 85.60      & 84.01  & \textbf{84.80}  \\ \hline
\multirow{6}{*}{\makecell{Twitter \\ (Ptá\v cek et al., 2014)}} 
& CNN-LSTM-DNN & 80.34          & 77.27     & 82.35  & 79.73 \\
& NBOW         & 75.18          & 75.25     & 75.18  & 75.18 \\
                                               & CNN          & 80.04          & 80.00        & 80.11  & 80.01 \\
                                               & SAWS         & 80.84          & 80.78     & 80.74  & 80.76 \\
                                               & ELMo         & 85.08          & 82.71     & 57.41  & 84.45 \\
                                               & Our model    & 93.63          & 92.51     & 94.05  & \textbf{93.27} \\ \hline
\end{tabular}
\end{table}

For each dataset, we trained each baseline model and our model. We observed that the basic models (i.e., CNN-LSTM-DNN, NBOW, and CNN) performed poorly compared to other models. These models simply encode input text to capture local semantic information, which is insufficient to derive global context or recognize incongruity between words. 

However, out of three above-mentioned models, CNN showed higher results on SARC/movies and SARC/technology subreddit. One of the reasons leading to such a result could be the relative shortness of texts in those datasets. The statistics are presented in  Table~\ref{tab:data stats}. Particularly,  median and mean for SARC/movies dataset $(Median = 10, Mean = 12.24)$ are significantly lower than median and mean for IAC-V2 dataset respectively $(Median = 39, Mean = 17.61)$.

\begin{table}[!htb]
\centering
\caption{Statistics of the datasets.}
\label{tab:data stats}
\begin{tabular}{|c|ccccc|}
\hline & \multicolumn{5}{c|}{Metrics}          \\ \cline{2-6} 
\multirow{-2}{*}{Datasets} & \multicolumn{1}{c|}{Median} & \multicolumn{1}{c|}{Mean}                          & \multicolumn{1}{c|}{Variance}                            & \multicolumn{1}{c|}{Max}                         & Min \\ \hline
SARC/movies               & \multicolumn{1}{c|}{10}     & \multicolumn{1}{c|}{12.24} & \multicolumn{1}{c|}{8.81} & \multicolumn{1}{c|}{138}                         & 1   \\
SARC/technology           & \multicolumn{1}{c|}{12}     & \multicolumn{1}{c|}{13.88} & \multicolumn{1}{c|}{9.33} & \multicolumn{1}{c|}{103}                         & 1   \\
IAC-V2                    & \multicolumn{1}{c|}{39}     & \multicolumn{1}{c|}{50.63} & \multicolumn{1}{c|}{36.05} & \multicolumn{1}{c|}{212} & 10  \\
\makecell{Twitter \\ (Ptá\v cek et al., 2014)}                    & \multicolumn{1}{c|}{17}     & \multicolumn{1}{c|}{17.61} & \multicolumn{1}{c|}{6.26} & \multicolumn{1}{c|}{64}                          & 1   \\ \hline
\end{tabular}
\end{table}

SAWS and ELMo models outperformed basic models. Compared to the previous baselines, SAWS and ELMo models are built to capture more sophisticated patterns, such as text fragments incongruity and complex morpho-syntactic features. 
 However, the ELMo model is almost always a few percents ahead of SAWS, showing that purely character-based input and subsequently obtained contextual embeddings capture more useful sarcastic information. 
 
For all datasets, our model outperformed all baselines for all metrics and achieved state-of-the-art performance. The best improvement of the F1 metric was achieved on IAC-V2 and Twitter datasets. Specifically, the F1 metric is  9\% higher, compared with the previous state-of-the-art version. For SARC/movies dataset, F1 metric is improved by 8\%, and for SARC/technology dataset F1 metric is improved by 2\%. 

%  We used different transformer models fitted on emotion detection and sentiment classification tasks to obtain contextualized and more deep features. This allows the model to learn complex patterns from different perspectives, e.g. ``emotional’’ and ``sentiment’’.
 
%  Furthermore, the CNN block of our model utilizes Glove embeddings in order to capture semantic relationships of words and general context.  Our model shows that modeling  dependencies between  emotion, sentiment, and sarcasm is an important feature for sarcasm detection task. 
 
Interestingly, all models show their best results on Twitter dataset, and their performance decrease when the length of the input text is relatively long (IAC-V2) or short (SARC/movies). It suggests that more ideas should be investigated for texts of a particular length.

\section{Ablation Study}

To understand the contribution of each module in our architecture, and thus the role of derived features in sarcasm detection task, we performed an ablation study. An ablation study examines the performance of a system by removing specific components in order to determine how they affect the system as a whole. 
% However, instead of removing one block of the architecture at a time, we replaced it with a simpler NBOW network. 

% NBOW network takes an average of word vectors in the given text as sentence representation and feeds it into a two-layers linear model. As word vector representations we used GloVe 300-dimensional vectors~\citep{pennington2014glove}.

The experiment was performed as follows: for each dataset, every individual module of the architecture was removed, and the new ``cut'' model was trained. Results of the experiments are presented in Table~\ref{tab:ablation_study}. Results of the integral model are included to make a comparison. In the table, similar results were highlighted in bald for F1 metric. 

\begin{table}[!htb]
\caption{Results of ablation study.}
\label{tab:ablation_study}
\begin{tabular}{|c|c|c|c|c|c|}
\hline
Dataset                                        & Removed Module & Acc.           & Precision & Recall & F1    \\ \hline
\multirow{5}{*}{SARC/movies}                   & SarcPTT        & 60.31          & 60.31     & 55.23  & 57.65 \\
                                               & EmoDPTT        & 70.38          & 69.08     & 71.45  & 70.25 \\
                                               & SentAPTT       & 73.36          & 71.92     & 74.73  & \textbf{73.30} \\ 
                                               % \cline{2-2}
                                               & CNN            & 72.9           & 71.73     & 73.63  & 72.67 \\
                                               & Integral Model & 73.36          & 72.05     & 74.41  & \textbf{73.21} \\ \hline
\multirow{5}{*}{SARC/technology}               & SarcPTT        & 61.72          & 64.13     & 80.8   & 71.5  \\
                                               & EmoDPTT        & 74.45          & 77.34     & 80.65  & 78.96 \\
                                               & SentAPTT       & 74.54          & 76.84     & 81.83  & 79.26 \\
                                               & CNN            & 73.57          & 75.41     & 82.42  & 78.76 \\
                                               & Integral Model & 75.24          & 76.73     & 83.75  & \textbf{80.08} \\ \hline
\multirow{5}{*}{IAC\_V2}                       & SarcPTT        & 65.72          & 64.72     & 67.24  & 65.96 \\
                                               & EmoDPTT        & 85.35          & 88.32     & 81.06  & \textbf{84.53} \\
                                               & SentAPTT       & 74.54          & 76.84     & 81.83  & 79.26 \\
                                               & CNN            & 84.13          & 87.48     & 79.19  & 83.13 \\
                                               & Integral Model & 85.12 & 85.6      & 84.01  & \textbf{84.80} \\ \hline
\multirow{5}{*}{Twitter (Ptacek et al., 2014)} & SarcPTT        & 78.15          & 75.07     & 80.06  & 77.48 \\
                                               & EmoDPTT        & 93.86          & 93.24     & 93.73  & \textbf{93.48}\\
                                               & SentAPTT       & 93.39          & 92.79     & 93.18  & 92.99 \\
                                               & CNN            & 93.46          & 92.74     & 93.38  & \textbf{93.06}\\
                                               & Integral Model & 93.63          & 92.51     & 94.05  & \textbf{93.27} \\ \hline
\end{tabular}
\end{table}

From the obtained results we can conclude, that performance of the model significantly decreased for all datasets when SarcPTT module was removed. Removing other modules affected the performance less than SarcPTT removal.

In case of SARC/movies dataset, removing EmoDPTT module decreased performance by 3\%. However, without SentAPTT block the performance was not affected.

For SARC/technology dataset the integral model outperformed all the ``cut'' models. 

For IAC\_V2 dataset removing SarcPTT and SentAPTT modules resulted in 19\% and 5\% F1 drop respectively. On the other hand, removing EmoDPTT did not affect the performance.

In case of Twitter dataset, removing SarcPTT decreased performance by 16\%. When EmoDPTT and CNN blocks were removed, the results were at par with the integral model.

SarcPTT block is based on RoBERTa architecture, which learns context features. Based on the observed results, we can conclude that context, even within a sentence or two, is the most important feature for sarcasm detection. Other blocks boost performance in some datasets more than others.

\section{Conclusion and Future Work}
 In this paper, we introduced a novel sarcasm detection method, which incorporates both emotion and sentiment features. Proposed approach contains four components: Sarcasm Pre-Trained Transformer (SarcPTT), Emotion Detection Pre-Trained Transformer (datasets), Sentiment Analysis Pre-Trained Transformer (SentAPTT), and CNN block. 
 Experiments were conducted on four datasets: SARC/movies, SARC/technology, IAC-V2, and Twitter. Our model showed significant improvement over the state-of-the-art models using all evaluation metrics. 

We performed the ablation study to measure the impact of each component in architecture on the final results. We discovered that the key performance contributor is SarcPTT block, which is based on RoBERTa model, pretrained on merged train chunks of all datasets.  Other modules provide features that improve performance in specific datasets.

% We discovered that emotion and sentiment features, as well as context features obtained by CNN block, are not as beneficial as context features in our architecture. The key contributor is SarcPTT block, which is based on RoBERTa model, pretrained on merged train chunks of all datasets. Thus, context plays a crucial role in the performance of sarcasm detection. 

 % EmoDPTT and SentAPTT are used as feature extractors and are not trainable during model fitting. These models are used to highlight the parts of the sentence which provide crucial cues for sarcasm detection. CNN module extracts semantic relationship of words and general context features. SarcPTT is pre-trained on train chunks of datasets, so the model learns sarcasm patterns. 
 %  The output from each block is passed through a fully-connected network and then through a linear layer to get the final classification score.

 Even though our model outperformed all the baselines, all the methods are struggling with too short or too long texts. 
  In our future works, we are planning to address this challenge. 

\section*{Acknowledgements}
The work was done with partial support from the Mexican Government through the grant A1-S-47854 of CONACYT, Mexico, grants 20220852 and 20220859 of the Secretaría de Investigación y Posgrado of the Instituto Politécnico Nacional, Mexico. The authors thank the CONACYT for the computing resources brought to them through the Plataforma de Aprendizaje Profundo para Tecnologías del Lenguaje of the Laboratorio de Supercómputo of the INAOE, Mexico, and acknowledge the support of Microsoft through the Microsoft Latin America Ph.D. Award.

\bibliography{bibliography}

\end{document}